\documentclass{article}

\usepackage[main, final]{neurips_2026}
\usepackage{amssymb}
\usepackage{marvosym}
\usepackage{fontawesome5}

\usepackage[utf8]{inputenc} 
\usepackage[T1]{fontenc}    
\usepackage{xcolor}
\definecolor{citeblue}{rgb}{0.21,0.49,0.74}
\usepackage[pagebackref,breaklinks,colorlinks,allcolors=citeblue]{hyperref}

\usepackage{url}            
\usepackage{booktabs}       
\usepackage{amsfonts}       
\usepackage{nicefrac}       
\usepackage{microtype}      
\usepackage{xcolor}         
\usepackage{amsmath} 
\usepackage{multirow}
\usepackage{graphicx}
\usepackage[table]{xcolor}

\newcommand{\ql}[1]{\textcolor{black}{#1}}
\newcommand{\dz}[1]{\textcolor{black}{#1}}

\newcommand{\lx}[1]{\textcolor{black}{#1}}

\title{PanoWorld: Real-World Panoramic Generation}

\author{%
  Haoyuan Li\textsuperscript{1} \quad 
  Dizhe Zhang\textsuperscript{1 $\text{\faEnvelope}$ }\thanks{Project Lead   \faEnvelope\ Corresponding Author}  \quad 
  Yuemei Zhou\textsuperscript{1}
  \quad 
  Xiangkai Zhang\textsuperscript{1, 2} \quad 
  Haoran Feng\textsuperscript{1, 3} 
  \AND
  Xiaofan Lin\textsuperscript{1} \quad 
  Wenjie Jiang\textsuperscript{1} \quad 
  Bo Du\textsuperscript{4} \quad 
  Ming-Hsuan Yang\textsuperscript{5} \quad 
  Lu Qi\textsuperscript{1,4 $\text{\faEnvelope}$} \\
  \textsuperscript{1}Insta360 Research \quad 
  \textsuperscript{2}Institute of Automation Chinese Academy
of Sciences\\
  \textsuperscript{3}Tsinghua University \quad 
  \textsuperscript{4}Wuhan University \quad
  \textsuperscript{5}UC Merced
}

\begin{document}

\maketitle

\maketitle
\begin{figure}[!h]
    \centering
    \includegraphics[width=\linewidth]{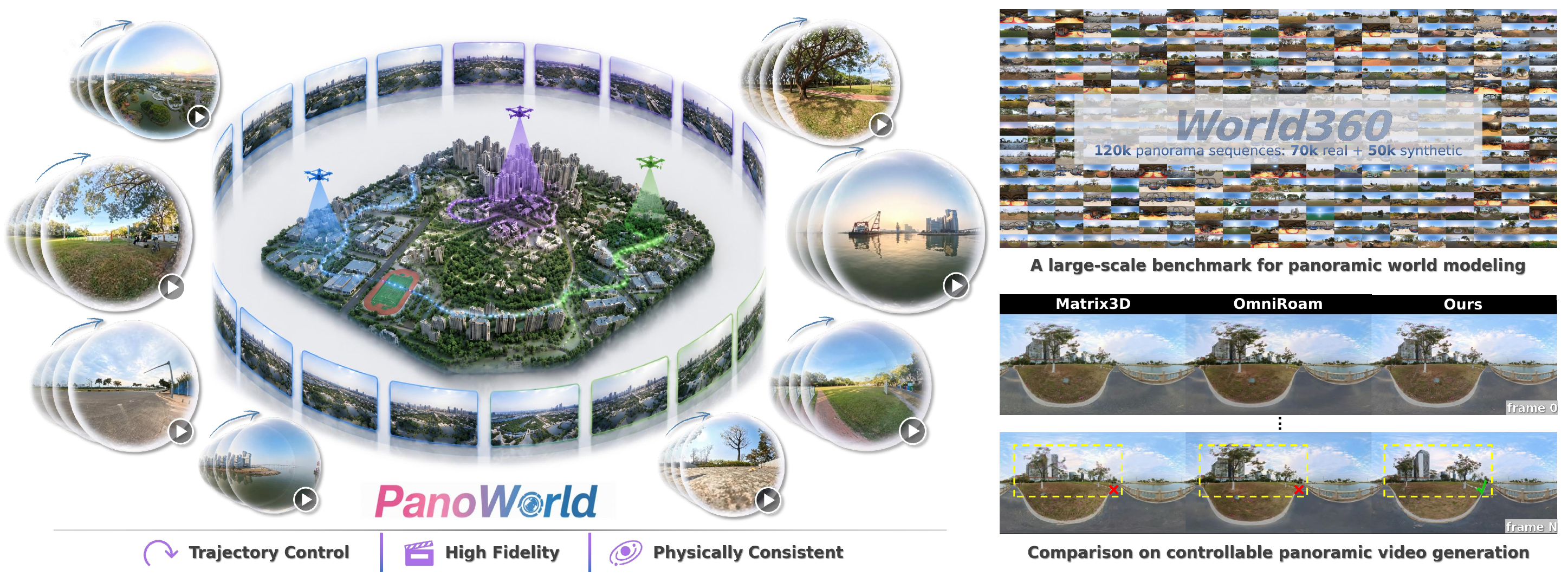}
    \caption{\textbf{PanoWorld} is a novel framework for high-fidelity and controllable panoramic video generation. Our approach achieves precise trajectory control across complex movements while maintaining high-fidelity visual synthesis with physical consistency in diverse real-world environments.
    }
    \label{fig: teaser}
\end{figure}

\begin{abstract}
\ql{
In this work, we aim to address the challenge of long-range memory in panoramic world models by exploiting the rotation-equivariant property of omnidirectional representations, where rotation can be treated as an implicit geometric transformation.
Building on this insight, we propose PanoWorld, which simplifies camera trajectories into translations via fixed headings for both current-action modeling and long-range memory through Dense Panoramic Ray-Conditioning (DPRC) and Geometry-aware Memory Augmentation (GMA).
Then, a three-stage training pipeline is introduced to progressively optimize each component.
To better evaluate physical consistency under large-scale spatial variations and diverse illumination conditions, where existing datasets are relatively stable, we construct World360, a large-scale dataset consisting of both real-world video clips collected via panoramic unmanned aerial vehicles and high-quality simulated clips generated by AirSim360.
Extensive experiments on World360 demonstrate the effectiveness of PanoWorld, outperforming alternative methods by a large margin.
Our models, training code, and dataset will be publicly available.  More information can be found on our project
page: https://lihaoy-ux.github.io/panoworld-page/.
}
\end{abstract}

\section{Introduction}
\ql{Recently, world models have attracted significant attention for modeling dynamic environments and enabling controllable generation ~\citep{tang2025hunyuan, team2026advancing}. They have shown strong potential in various robotic applications, including autonomous driving and unmanned aerial vehicles ~\citep{wu2026pragmatic}.}

\ql{Among various world models, panoramic representations, particularly equirectangular projections (ERP), have emerged as a mainstream paradigm by capturing the full $360^\circ$ field of view (FoV) for each frame in a specific trajectory ~\citep{lin2025one}. Despite recent advancements, achieving physical consistency, such as geometry and illumination, across space and time remains challenging for panoramic world models, where such problem can be amplified by the full-view nature.}


\ql{
Most existing work, including 3DGS-based and video-generation paradigms, adopts memory mechanisms (e.g., 3D points or KV caches) to retrieve past information for spatiotemporal consistency ~\citep{yang2025matrix, chou2025captain, schwarz2025generative}.
However, these solutions inherit perspective assumptions and ignore the unique properties of panoramic data, leading to misaligned memory retrieval under severe distortion and rotation-induced viewpoint shifts.
Thus, one question raised: how can memory mechanisms better adapt panoramic representations?
}

\ql{To address this issue, we begin by analyzing the properties of the equirectangular projection (ERP), which is rotation-equivariant, where rotations mainly alter the distortion pattern while preserving the underlying scene content.
This observation inspires us to simplify camera motion by treating rotation as an geometric transformation and modeling only translation explicitly.
Building on this insight, we propose PanoWorld, a diffusion-based framework that incorporates Dense Panoramic Ray-Conditioning (DPRC) and Geometry-aware Memory Augmentation (GMA) modules for both current-action modeling and long-range memory, while decoupling translation and rotation via fixed headings.
Then, a three-stage training pipeline is introduced to effectively optimize each component.
}

\ql{
Given that existing datasets are predominantly indoor or simulation-based, where physical conditions are relatively stable, they are less suitable for training models and evaluating real-world physical consistency.
We further construct a large-scale dataset, World360, comprising 70K real-world clips collected via Anti-Gravity and 50K high-fidelity simulations from AirSim360 platform~\citep{ge2025airsim360}.
And the extensive experiments on World360 demonstrate the effectiveness of our method, achieving a large margin over alternative approaches.
}

\ql{Our contributions are summarized as follows:}
\begin{itemize}
    \item \ql{We propose PanoWorld, a diffusion-based panoramic world model that addresses long-range memory by leveraging the rotation-equivariant property of omnidirectional representations.}
    \item \ql{The PanoWorld consists of two core modules, Dense Panoramic Ray-Conditioning (DPRC) and Geometry-aware Memory Augmentation (GMA), together with a three-stage training pipeline. DPRC provides dense geometric conditioning for panoramic motion modeling, while GMA enhances memory retrieval and spatiotemporal consistency in panoramic space.}
    \item \ql{We construct World360, a large-scale benchmark for panoramic world modeling under real-world physical variations. Extensive experiments demonstrate that PanoWorld consistently outperforms existing alternatives by a large margin.}
    
\end{itemize}

\section{Related Work }
\paragraph{Panoramic Video Generation.} 

Panoramic vision provides a holistic geometric perspective for globally coherent scene synthesis. Recent works such as WorldPrompter~\citep{zhang2025generating} demonstrate that panoramic generation can serve as an effective intermediate representation for improving structural consistency. Early foundational frameworks, including 360DVD~\citep{wang2024360dvd}, pioneered the text-to-panoramic video generation pipeline and introduced the WEB360 dataset for benchmarking. To improve data scale and generation quality, PanoWan~\citep{xia2025panowan} builds upon large-scale immersive datasets, 360-1M~\citep{wallingford2024image}. Meanwhile, recent methods, including 4K4DGen~\citep{xing2025tip4gen} and DynamicScaler~\citep{liu2025dynamicscaler} focus on enhancing spatiotemporal and geometric consistency in panoramic video generation.
However, despite these advances, existing panoramic video generators still struggle in complex outdoor environments, where maintaining structural and radiometric consistency remains challenging.

\paragraph{Camera-controlled Video Diffusion Models. }
Controlling camera motion during video generation has become a key direction for decoupling scene dynamics from scene~\citep{bai2025recammaster, wang2024motionctrl, zhang2026unified}. Recent methods such as CameraCtrl~\citep{he2024cameractrl} and ViewCrafter~\citep{yu2024viewcrafter} incorporate extrinsic parameters or point cloud-based guidance to enable controllable generation. In panoramic settings, 360DVD~\citep{wang2024360dvd} introduces spherical motion conditioning, though accurate 3D trajectory control remains challenging; Matrix-3D~\citep{yang2025matrix} reconstructs scene geometry from an initial frame and performs 3D-to-2D rendering with mask-based inpainting, while OmniRoam~\citep{liu2026omniroam} adopts a ReCamMaster-style pipeline ~\citep{bai2025recammaster} that combines real and generated frames for consistent view synthesis. However, these methods often rely on explicit reconstruction or frame concatenation, which can introduce artifacts, increase computational cost, and limit scalability. 
\paragraph{Memory-Augmented Video Synthesis.}
A fundamental challenge in long-horizon video generation is scene drift due to the lack of persistent memory~\citep{song2025history}. Common baselines like Context-as-Memory~\citep{yu2025context} remain clip-bound and fail to capture long-range dependencies. While methods like Captain Safari~\citep{chou2025captain} use implicit 3D features to improve consistency, they require full-sequence processing and are unsuitable for open-world generation. For panoramic video, existing approaches such as Matrix-3D~\citep{yang2025matrix} rely on expensive explicit 3D reconstruction, which limits scalability and hinders real-time deployment. In contrast, we propose a framework that leverages the inherent rotation-equivariance of panoramic geometry to achieve long-range consistency and precise control without the need for exhaustive global optimization.

\begin{figure}[t] 
  \centering
  \includegraphics[width=0.95\textwidth]{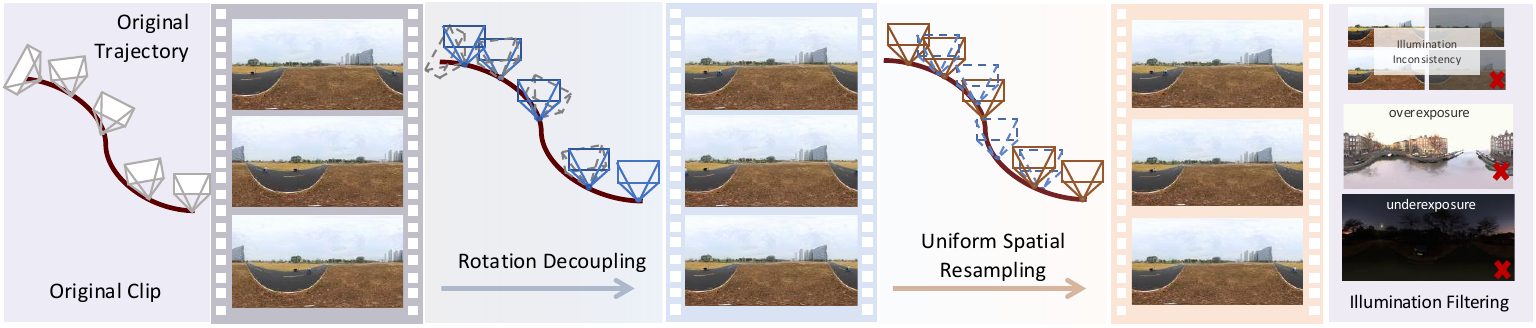} 
  
  \caption{\textbf{The Data Curation Pipeline.} The pipeline converts raw panoramic clips into high-quality sequences by: (1) \textit{Rotation Decoupling} to isolate pure translation; (2) \textit{Uniform Spatial Resampling} to standardize motion scales; and (3) \textit{Illumination Filtering} to ensure exposure consistency.}
  \label{data-curation} 
  \vspace{-0.2in}
\end{figure}

\section{Data Pipeline and Benchmark}

\dz{In this section, we detail our specialized data engine designed to facilitate large-scale controllable panoramic video synthesis. We first describe our curation pipeline which enforces physically-grounded consistency across temporal, geometric, and illumination dimensions. We then introduce World360, a large-scale benchmark specilized for panoramic world modeling under real-world scenarios.}

\paragraph{Data Curation}

\dz{To anchor scene evolution directly to camera movement, our engine operates through a three-fold alignment strategy (as shown in Fig.\ref{data-curation}): \textbf{Rotation Decoupling} rectifies panoramic sequences by factoring out relative rotations and re-projecting frames into a unified heading. By isolating translation as the primary driver of visual change, this process simplifies the learning of spatial depth and motion parallax within an panoramic ray-conditioned manifold. \textbf{Uniform Spatial Resampling} then replaces traditional fixed-time sampling with a constant spatial increment ($\Delta d$), eliminating velocity variations to establish a uniform geometric scale across all trajectories. Finally, \textbf{Illumination Filtering} audits the data to exclude sequences with improper illumination conditions, ensuring the model learns a persistent radiometric representation. Together, these processes eliminate geometric and photometric noise, providing a foundation for physically-consistent world modeling.}

\paragraph{World360}
\dz{Based on the aforementioned pipeline, we construct World360, a comprehensive benchmark for panoramic world modeling under real-world physical variations. As summarized in Table~\ref{tab:dataset_comparison}, World360 comprises 120,000 high-quality sequences, unifying 70,000 curated real-world clips with 50,000 high-fidelity simulations from AirSim360~\citep{ge2025airsim360}.}
Unlike datasets restricted to planar motion, World360 introduces diverse multi-altitude aerial trajectories with precise camera poses and depth information. This benchmark bridges the gap between constrained modeling and unconstrained panoramic synthesis, enabling the evaluation of structural integrity across complex 3D paths.

\begin{table}[htbp]
\centering
\caption{Comparison of panorama datasets for world generation.}
\label{tab:dataset_comparison}
\resizebox{\textwidth}{!}{
\begin{tabular}{lcccccccc}
\toprule
\textbf{Dataset} & \textbf{Source} & \textbf{\#Samples} & \textbf{Motion Space} & \textbf{Pose} & \textbf{Depth} & \textbf{Text} & \textbf{Res.} & \textbf{Video} \\ \midrule
Imagine360 [46]  & Real  & 10k    & Planar          & no  & no  & \checkmark & Mixed & \checkmark \\
Web360 [53]      & Real  & 2k     & Planar          & no  & no  & \checkmark & 720p  & \checkmark \\
Argus [34]       & Real  & 283k   & Planar          & no  & no  & no  & Mixed & \checkmark \\
360-1M [49]      & Real  & 1,076k & Planar$^\ddagger$ & \checkmark & no  & no  & Mixed & \checkmark \\
PanoWan [56]     & Real  & 13k    & Planar          & no  & no  & \checkmark & Mixed & \checkmark \\
Matrix-Pano      & Syn.  & 116k   & 3D Space        & \checkmark & \checkmark & \checkmark & 2K    & \checkmark \\ \midrule
\textbf{World360 (Ours)} & \textbf{Real} & \textbf{120k} & \textbf{Multi-Alt.} & \checkmark & \textbf{w/ Syn.} & \checkmark & \textbf{2K} & \checkmark \\ \bottomrule
\end{tabular}
}
\begin{flushleft}
\footnotesize $^\ddagger$ Existing large-scale real datasets primarily consist of \textbf{planar street-level motion}, where altitude information is ambiguous or fixed. In contrast, \textbf{World360} provides diverse \textbf{aerial trajectories} across multiple altitudes.
\end{flushleft}
\vspace{-10pt}
\end{table}

\section{Our Method}

\lx{
Given an initial panoramic observation and a target camera trajectory, our framework first removes rotational ambiguity to simplify the motion space, then injects geometry-aware motion control and historical memory into a unified diffusion backbone, and finally decodes the fused representation into a controllable panoramic video. 
This section is organized into two parts: Sec. \ref{4.1} presents the core modeling design underlying motion control and memory modeling, and Sec. \ref{4.2} describes the three-stage optimization strategy used to train the full framework.
}

\subsection{Model Design}
\label{4.1}

\begin{figure}[t] 
  \centering
  \includegraphics[width=0.95\textwidth]{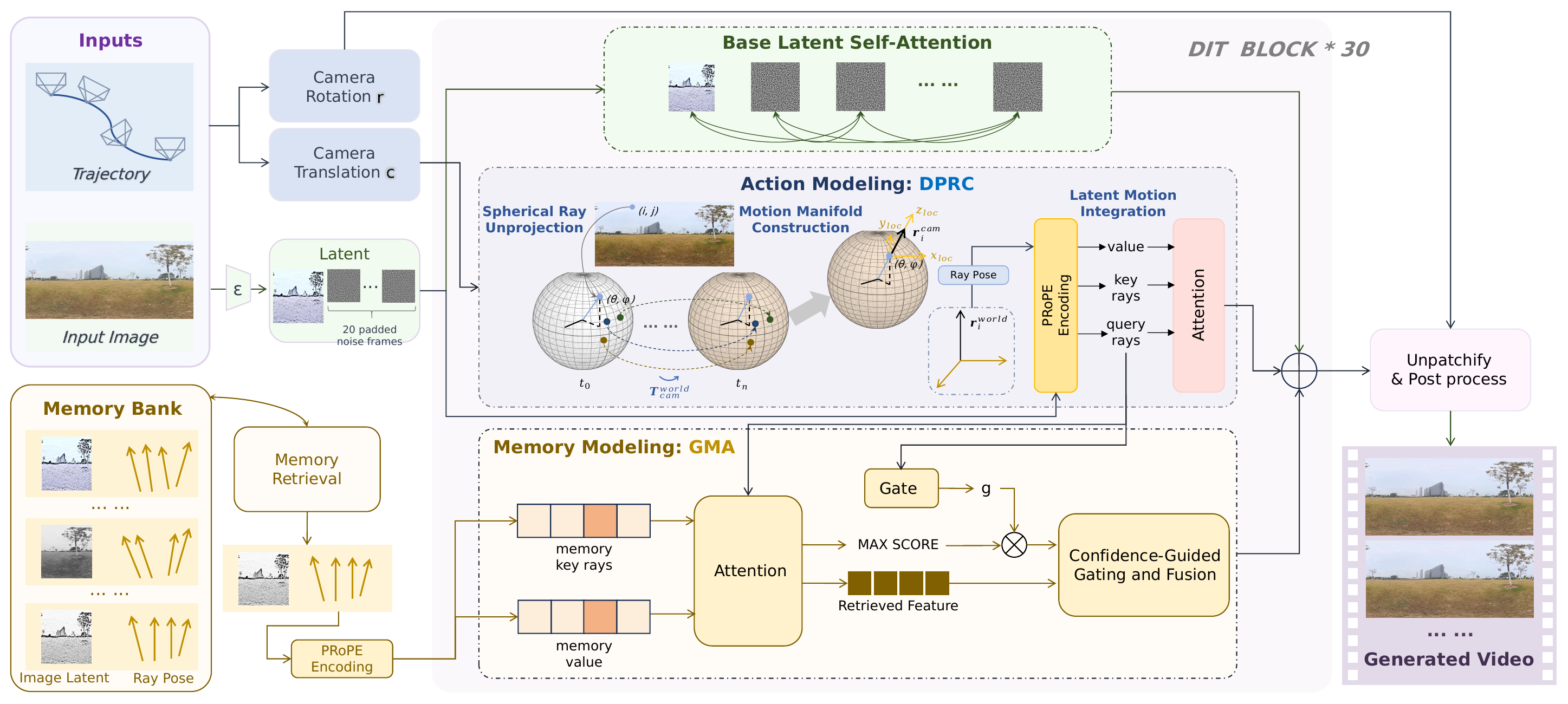} 
  \caption{\textbf{PanoWorld Network Architecture.} Built on the Wan2.2 backbone, PanoWorld employs a triple-stream DiT that fuses visual self-attention, DPRC-based action modeling, and a GMA module for memory-anchored synthesis via a shared geometric manifold.
  }
  \label{data-engine} 
\end{figure}

\lx{
This section presents three core modeling designs that make the generation physically consistent, as illustrated in Fig.~\ref{data-engine}. 
We first simplify the learning problem through motion decoupling (\ref{4.1.1}), then model view-dependent translation with DPRC (\ref{4.1.2}), and finally enforce long-horizon scene persistence through geometry-aware memory modeling (\ref{4.1.3}).
}

\subsubsection{Motion Decoupling Principle}

\label{4.1.1}

Conventional models often learn translation and rotation jointly, which lead to structural warping in panoramic video generation. 
To maintain geometric consistency, we observe that camera rotation in a panoramic sequence is independent of scene depth, and therefore isolate it as a precise geometric transformation, allowing the diffusion model to focus exclusively on translation-induced parallax. 
To bridge the gap between pre-trained perspective priors and panoramic geometry, we further fine-tune the backbone on panoramic datasets with LoRA, enabling it to capture intrinsic equirectangular properties such as polar distortion and horizontal continuity while preserving the generative quality of the base model. 
Combined with explicit motion decoupling, this adaptation removes artifacts that violate $360^\circ$ projection rules and preserves structural integrity across camera movements.

\subsubsection{Action Modeling: DPRC}

\label{4.1.2}
To achieve unconstrained motion control while enforcing joint consistency across temporal, geometric, and radiometric domains, we propose the Dense Panoramic Ray-Conditioning (DPRC) mechanism. Unlike traditional perspective models that rely on pixel-level optical flow, DPRC leverages equirectangular geometry to model the generative process as a dynamic light-field evolution, aligning kinematic priors with the physical origin of panoramic rays

\paragraph{Spherical Ray Unprojection.} We first map each pixel $(i, j)$ of the latent grid to a unit ray direction $\mathbf{r}_{cam} \in \mathbb{S}^2$. Given the dimensions as height $H$ and width $W$, the latitude $\theta$ and longitude $\phi$ are:\begin{equation}\theta = \frac{\pi}{2} - \frac{i + 0.5}{H} \pi, \quad \phi = \frac{j + 0.5}{W} 2\pi - \pi\end{equation}Following our dataset convention, the resulting camera-space ray is defined as:
\begin{equation}\mathbf{r}_{cam} = [-\cos\theta \cos\phi, -\cos\theta \sin\phi, -\sin\theta]^\top
\end{equation}

This mapping ensures subsequent motion modeling respects the scene’s inherent $360^\circ$ topology, mitigating non-uniform planar projection distortion.
\paragraph{Motion Manifold Construction.} 
To maintain radiometric consistency during 3D translation, the model must perceive how ray intensity evolves with camera displacement.
Since we employ a de-routed representation (where $\mathbf{R}_t = \mathbf{I}$), the motion manifold is determined solely by the translation $\mathbf{c}_t$. We construct this manifold through a three-step local transformation:

Step 1: Local Orthonormal Basis. For each static ray, we construct a right-handed orthonormal basis $\mathbf{R}_{loc} = [\mathbf{a}_x, \mathbf{a}_y, \mathbf{z}_{loc}]$, where the local z-axis aligns with the ray direction: $\mathbf{z}_{loc} = \mathbf{r}_{cam}$. \begin{equation}\mathbf{a}_y = \text{normalize}(\mathbf{z}_{loc} \times \mathbf{u}_{world}), \quad \mathbf{a}x = \mathbf{a}y \times \mathbf{z}_{loc}
\end{equation}
where $\mathbf{u}_{world} = [0, 0, -1]^\top$ is the world-up vector.

Step 2: Local Transformation Matrix. We define a matrix $\mathbf{T}_{ray}$ that anchors the global camera motion to the individual ray's perspective:
\begin{equation}
\mathbf{T}_{\text{ray}} = \begin{bmatrix}
\mathbf{R}_{\text{loc}}^\top & -\mathbf{R}_{\text{loc}}^\top \mathbf{c}_t \\
\mathbf{0}^\top & 1
\end{bmatrix} \in SE(3)
\end{equation}
The translation term $-\mathbf{R}_{loc}^\top \mathbf{c}_t$ represents the projection of the camera center $\mathbf{c}_t$ onto the ray's local axes. This explicit encoding of translational parallax informs the model how the camera's displacement affects the observed scene from that specific ray's vantage point.

\paragraph{Latent Motion Integration.} 
To ensure computational efficiency, $\mathbf{\hat{r}}_{t}$ is encoded via Projective Positional Embeddings (PRoPE) and injected directly into the diffusion transformer blocks. By operating on the downsampled latent grid, the model maintains precise kinematic control with minimal overhead. This high-fidelity integration ensures temporal and radiometric consistency, as the generated visual flow is constrained by the inherent geometry of the ray-field, preventing structural drift and ensuring view-dependent stability across complex maneuvers.

\subsubsection{Memory Modeling: GMA}
\label{4.1.3}
To ensure robust spatio-temporal and radiometric consistency across evolving trajectories, we propose the Geometry-Aware Memory (GMA) mechanism. Unlike methods relying on explicit 3D primitives, GMA facilitates historical content retrieval through an implicit geometric manifold, anchoring the generative process to a persistent 3D world.

\paragraph{Unified Projection for Query and Memory.}GMA unifies motion conditioning and memory retrieval within a shared geometric coordinate system by encoding query latents and memory bank features $\{\mathbf{x}_m\}$ into a common PRoPE space. Memory keys and values are formulated as :
\begin{equation}[\mathbf{K}_{mem},\mathbf{V}_{mem}] = \text{PRoPE}(\text{Linear}(\mathbf{x}_m), \mathbf{\hat{r}}_{m}),
\end{equation}
where $\mathbf{\hat{r}}$ is the same ray-based representation used for motion control. This alignment enables the attention mechanism to retrieve historical features based on 3D ray correspondence rather than explicit image-space warping. Consequently, when the camera revisits locations, retrieved features maintain inherent geometric and radiometric consistency with original observations. 
\paragraph{Confidence-Guided Gating and Fusion.}To prevent content hallucination in unobserved regions and maintain global stability, we implement a Confidence-Guided Gating mechanism. We quantify retrieval confidence $c$ based on the maximal attention affinity, identifying regions with high geometric overlap:
\begin{equation}c = \text{clamp}\left( \max_{j} (\text{Softmax}(\mathbf{Q}_i \mathbf{K}_j^\top / \sqrt{d})), 0, 1 \right)
\end{equation}
The final feature $\mathbf{F}_{final}$ is synthesized by fusing the retrieved memory $\mathbf{F}_{mem}$ with the base diffusion feature $\mathbf{F}_{base}$ via an adaptive gate:
\begin{equation}\mathbf{F}_{final} = \mathbf{F}_{base} + (g \cdot c) \cdot \mathbf{F}_{mem}
\end{equation}
The final feature $\mathbf{F}_{final}$ is synthesized through an adaptive fusion: $\mathbf{F}_{base} + (g \cdot c) \cdot \mathbf{F}_{mem}$. The result is a persistent generative framework that enforces long-term radiometric consistency. By anchoring synthesis to high-confidence historical rays, the model eliminates flickering and structural drift during complex maneuvers, ensuring that the visual properties of the environment remain invariant across the entire trajectory.

\begin{figure}[t] 
  \centering
  \includegraphics[width=0.95\textwidth]{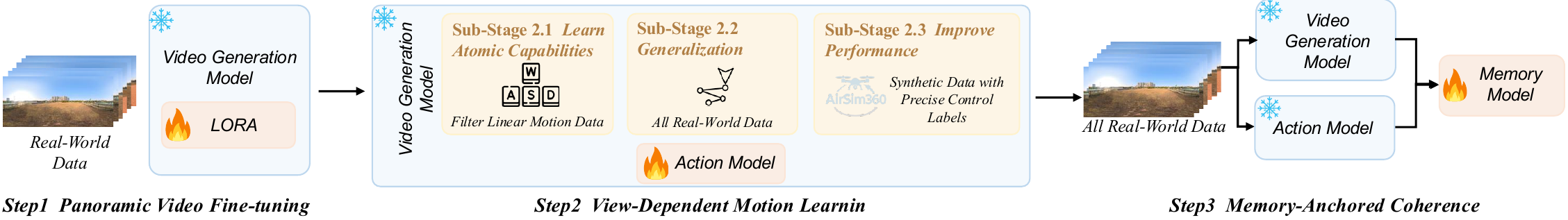} 
  \caption{Overview of the progressive three-stage pipeline. \textit{Stage 1} focuses on geometric adaptation for panoramic video generation; \textit{Stage 2} targets view-dependent motion learning ; and \textit{Stage 3} enables memory-anchored coherence to enforce long-term temporal consistency.}
  \label{training_pipeline} 
\vspace{-0.2in}
\end{figure}

\subsection{Training Strategy}
\lx{
This section presents how the above components are optimized progressively rather than all at once, as illustrated in Fig.~\ref{training_pipeline}. 
We first adapt the backbone to panoramic geometry, then learn motion control, and finally activate memory modeling, which stabilizes optimization and helps each component acquire its intended role before full integration.
}
\label{4.2}

\paragraph{Stage 1: Panoramic Video Fine-tuning.}
The goal of this stage is to adapt the pre-trained video model to the panoramic domain. We use LoRA to fine-tune the backbone on panoramic datasets, allowing it to learn equirectangular geometric features such as polar distortions and horizontal wrap-around consistency. To balance the training, we use a Latitude-Aware Reconstruction Loss:
\begin{equation}
\mathcal{L}{stage1} = \mathbb{E} \left[ \cos(\theta) \cdot | \epsilon - \epsilon\theta(\mathbf{x}_t, t) |^2 \right]
\end{equation}
The result is a geometry-aware backbone that inherently respects $360^\circ$ topological constraints. This establishes a foundation for geometric consistency, preventing structural tearing and boundary artifacts during synthesis.

\paragraph{Stage 2: View-Dependent Motion Learning.}
In this stage, we freeze the LoRA weights and train the Action Stream via DPRC to master 3D translational control. By removing yaw rotation from training trajectories, we force the model to isolate depth-dependent parallax learning. DPRC models ego-motion as a dense re-indexing of panoramic rays, allowing the generator to learn intensity evolution relative to 3D spatial coordinates. This ray-geometry alignment enforces radiometric consistency, ensuring stable appearances and structural integrity during complex maneuvers.

\paragraph{Stage 3: Memory-Anchored Coherence.}
The final stage activates the GMA module, transitioning from sequential generation to memory-anchored synthesis via a unified geometric manifold. Using Confidence-Guided Gating, the model adaptively anchors synthesis to stored landmarks. This framework enforces long-term radiometric consistency by ensuring previously observed landmarks remain identical upon revisit. By anchoring synthesis to high-confidence historical rays, the model eliminates flickering and structural drift, providing a continuous and stable 3D scene realization.

\section{Experiments}
\subsection{Experiment Settings}
\paragraph{Implementation Detail}
We implement PanoWorld by fine-tuning the Wan2.2-5B \lx{architecture} architecture. To efficiently adapt the base model while preserving its generative priors, we employ LoRA-based fine-tuning. As described in Section 4.2, our training follows a decoupled, multi-stage pipeline \lx{in which} the action-conditioning and memory modules are optimized independently \lx{under} parameter freezing.
\paragraph{Evaluation Metrics}
We evaluate our framework using a multi-dimensional suite of metrics covering distribution fidelity, visual quality, and panoramic-specific perception. We report FID and its regional variants ($FID_{pole}$ \lx{and} $FID_{equ}$) to assess structural integrity across the omnidirectional manifold. Visual fidelity and temporal consistency are measured \lx{using} $FAED$, while the $NIQE$ and Q-Align metrics ($QA_{qual.}$ \lx{and} $QA_{aes.}$) ~\citep{wu2023q} quantify clarity and aesthetic coherence.
To evaluate \lx{trajectory controllability}, we follow the protocol of CamPVG ~\citep{ji2025campvg} and OminiRoam ~\citep{liu2026omniroam}. \lx{Specifically,} all methods are conditioned on the same ground-truth (GT) trajectories to avoid unfair \lx{comparisons} caused by varying motion scales or noise \lx{introduced} by structure-from-motion extraction. We compute the average PSNR between generated videos and GT sequences over multiple temporal windows to quantify adherence to the specified trajectory. Higher PSNR values indicate more accurate camera control and \lx{stronger} structural persistence over extended horizons.

\begin{figure}[t]
  \centering
  \includegraphics[width=\textwidth]{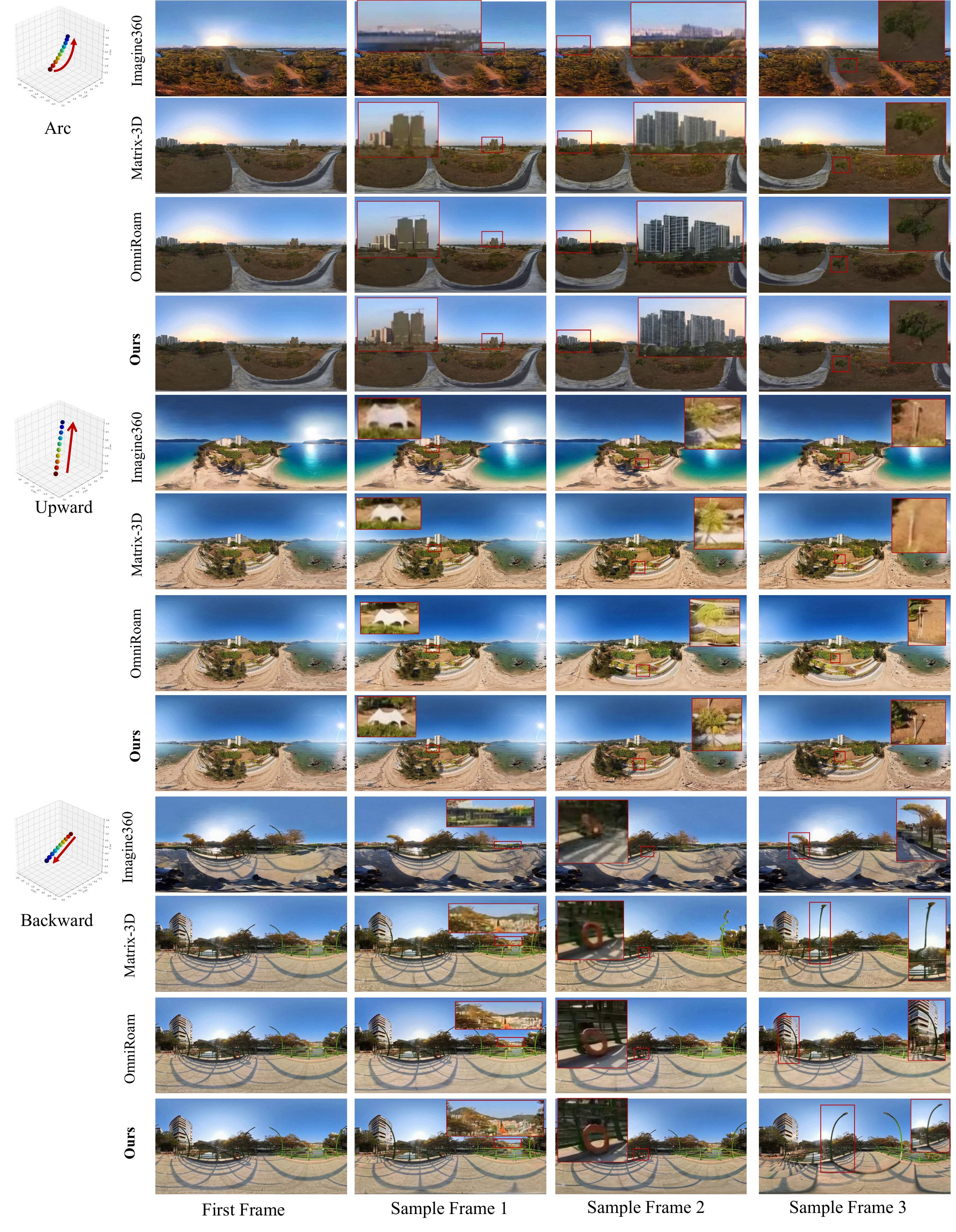} 
  \caption{Qualitative comparison on real-world outdoor sequences.}
  \label{resutls_img} 
\end{figure}

\paragraph{Baselines}
We evaluate the performance of our proposed method against three state-of-the-art frameworks for trajectory-controlled panoramic video synthesis: Imagine360 ~\citep{tan2024imagine360}, Matrix-3D ~\citep{yang2025matrix}, and OmniRoam ~\citep{liu2026omniroam}. For a fair comparison, all models are provided with a unified input resolution of 480p ($480 \times 960$), 720p ($720 \times 1440$) and \lx{de-rotated camera trajectories} derived from the same evaluation dataset. Specifically, since Imagine360 originally generates 32-frame videos at a $512 \times 1024$ resolution, we downsample its output to $480 \times 960$ and apply temporal interpolation to extend the sequence to 81 frames, \lx{ensuring consistency with the evaluation protocol}.
\subsection{Results and Comparisons}
\paragraph{Qualitative Results} Figure \ref{resutls_img} \lx{presents} qualitative comparisons between our method and existing approaches using real-world outdoor sequences. We visualize the generation results across diverse movement patterns to assess both visual fidelity and motion consistency.

Our method maintains precise trajectory adherence and geometric stability with superior semantic clarity. In contrast, Matrix-3D suffers from significant blurring and structural "voids" during height variations, despite its explicit 3D priors. OmniRoam fails to follow vertical instructions due to its training on fixed-height data, resulting in severe ghosting and artifacts. These results confirm that PanoWorld effectively mitigates structural drift and maintains environmental persistence, outperforming baselines in complex real-world maneuvers.

\paragraph{Quantitative Analysis of Visual Quality.}
As shown in Table \ref{tab:main_results}, PanoWorld outperforms baseline methods across nearly all visual quality metrics. In terms of distribution fidelity, our method achieves the lowest FID (27.64), FID$_{pole}$ (47.21), and FID$_{equ}$ (26.00). These results indicate that our model maintains superior structural integrity across the entire $360^{\circ}$ manifold, effectively suppressing the severe distortions at the poles and blurriness at the equator that often plague other panoramic models. Furthermore, our approach achieves a leading QA$_{qual.}$ (4.0202), reflecting high visual realism. While Imagine360 shows higher aesthetic scores, it lacks the geometric stability and physical consistency provided by our framework.

\begin{table}[t]
\centering
\caption{Quantitative results on panoramic video generation. Best results are in red.}
\label{tab:main_results}
\begin{small}
\begin{tabular}{l c c c c c c c c}
\toprule
\multirow{2}{*}{Method} & \multirow{2}{*}{Resolution} & \multicolumn{3}{c}{Distribution Fidelity $\downarrow$} & \multicolumn{1}{c}{Alignment $\downarrow$} & \multicolumn{3}{c}{Panoramic Perceptual Quality} \\
\cmidrule(r){3-5} \cmidrule(lr){6-6} \cmidrule(l){7-9}
&  & FID $\downarrow$ & FID$_{pole}$ $\downarrow$ & FID$_{equ}$ $\downarrow$ & FAED $\downarrow$ & NIQE $\downarrow$ & QA$_{qual.} \uparrow$ & QA$_{aes.} \uparrow$ \\
\midrule
Imagine360 & 480p & 81.18 & 113.95 & 60.99 & 4.50 & \cellcolor{red!25}3.15 & 4.0915 & \cellcolor{red!25}3.4657 \\
Matrix-3D  & 480p & 34.63 & 67.88 & 68.64 & 3.39 & 4.74 & 2.9942 & 2.1258 \\
OmniRoam   & 480p & 60.77 & 85.25 & 45.92 & 3.36 & 3.39 & 3.7510 & 3.0575 \\

\textbf{Ours} & 480p & \cellcolor{red!25}27.64 & \cellcolor{red!25}47.21 & \cellcolor{red!25}26.00 & \cellcolor{red!25}2.63 & 3.85 & \cellcolor{red!25}4.0202 & 3.1283 \\
\midrule
Matrix-3D  & 720p & 35.35 & 70.27 & 44.06 & 2.86 & 3.63 & 3.75 & 2.9653 \\
OmniRoam   & 720p & 44.89 & 89.95 & 83.24 & 3.06 & 3.74 & 4.07 & \cellcolor{red!25}3.5134 \\

\textbf{Ours} & 720p & \cellcolor{red!25}16.93 & \cellcolor{red!25}35.46 & \cellcolor{red!25}18.86 & \cellcolor{red!25}1.18 & \cellcolor{red!25}3.24 & \cellcolor{red!25}4.14 & 3.3413 \\
\bottomrule
\end{tabular}
\end{small}
\end{table}

\paragraph{Quantitative Analysis of Trajectory Control.}
Our framework consistently outperforms Matrix-3D and OmniRoam across all temporal windows. As shown in Table \ref{tab:motion_stability}, PanoWorld achieves significantly higher PSNR, reflecting superior path adherence. Notably, OmniRoam suffers from substantial structural drift during complex elevation changes in high-altitude scenarios. We further validated trajectory fidelity by using ViPE ~\citep{huang2025vipe} to reconstruct camera poses from the generated videos; our method consistently demonstrates the tightest alignment with ground-truth commands across both stability and pose-estimation metrics. These results confirm PanoWorld's robustness in executing precise maneuvers in challenging large-scale environments.

\begin{table}[t]
\centering
\caption{Motion control evaluation. Best results are in red, while `--` denotes metrics not supported by the baseline's original implementation.}
\label{tab:motion_stability}
\begin{small}
\begin{tabular}{l c cccc}

\toprule
Method & Resolution & PSNR$_{20-25}$ & PSNR$_{50-55}$ & PSNR$_{70-75}$ & PSNR$_{75-80}$ \\
\midrule
Imagine360 & 480p & -- & -- & -- & -- \\
Matrix-3D   & 480p & 20.47 $\pm$ 2.81 & 18.80 $\pm$ 2.67 & 18.13 $\pm$ 2.68 & 18.02 $\pm$ 2.68 \\
OmniRoam    & 480p & 18.51 $\pm$ 4.40 & 17.59 $\pm$ 4.47 & 17.34 $\pm$ 4.01 & 17.02 $\pm$ 3.83 \\
\textbf{Ours} & 480p& \cellcolor{red!25}22.83 $\pm$ 3.73 & \cellcolor{red!25}21.65 $\pm$ 3.70 & \cellcolor{red!25}21.04 $\pm$ 3.73 & \cellcolor{red!25}20.92 $\pm$ 3.73 \\
\midrule
Matrix-3D   & 720p & 20.30 $\pm$ 2.99 & 18.79 $\pm$ 2.89 & 18.26 $\pm$ 2.85 & 18.16 $\pm$ 2.83 \\
OmniRoam    & 720p & 18.63 $\pm$ 3.78 & 17.78 $\pm$ 3.60 & 17.37 $\pm$ 3.49 & 17.30 $\pm$ 3.44 \\
\textbf{Ours} & 720p& \cellcolor{red!25}22.94 $\pm$ 3.77 & \cellcolor{red!25}21.41 $\pm$ 3.75 & \cellcolor{red!25}20.83 $\pm$ 3.84 & \cellcolor{red!25}20.74 $\pm$ 3.87 \\
\bottomrule
\end{tabular}
\end{small}
\end{table}

\begin{figure}[htbp]
\centering
\includegraphics[width=\textwidth]{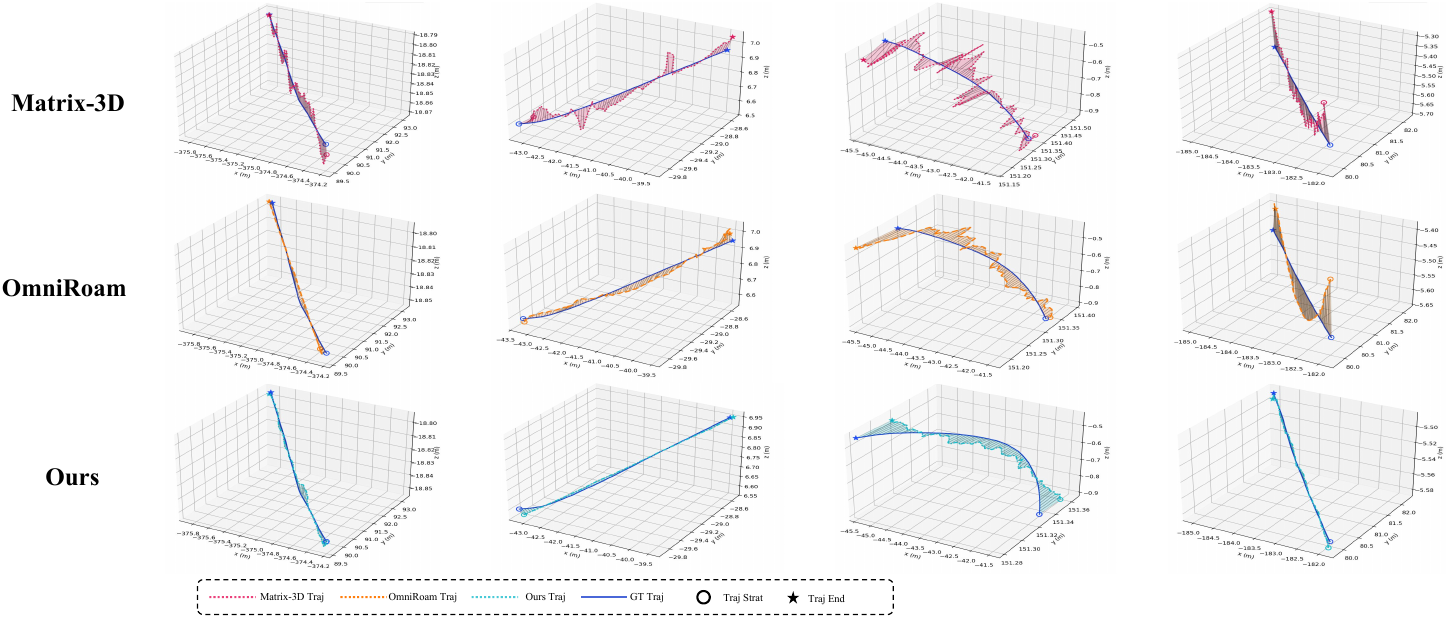}
\caption{{Qualitative Comparison Evaluation of Trajectory Fidelity via ViPE.} We utilize ViPE to reconstruct camera trajectories from the synthesized videos, comparing them against the GT. }
\label{results_vibe}
\end{figure}

\subsection{Ablative Studies}
As shown in Table \ref{tab:ablation_memory}, the full model is compared against w/o GMA and Random Memory variants, and GMA significantly improves geometric stability and long-term trajectory adherence, with performance gains scaling alongside sequence length.

\begin{figure}[htbp] 
  \centering
  \includegraphics[width=\textwidth]{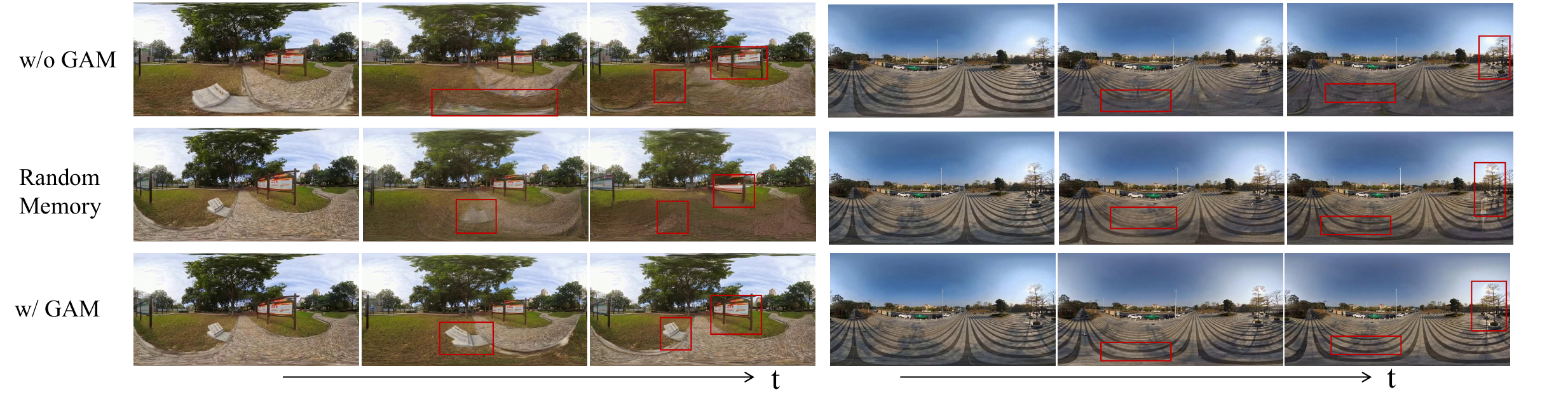} 
  \caption{Qualitative ablation results of the GMA module. Comparison of synthesized video frames under control: (a) w/o GMA baseline exhibits significant  artifacts; (b) Random Memory variant results in broken geometry; (c) Full setting with GMA maintains consistent generation.}
  \label{resutls} 
\end{figure}

\begin{table}[htbp]
\centering
\caption{Ablation study on the Memory Module. We compare trajectory control stability (PSNR) across different temporal windows. \textbf{Bold} indicates the best performance.}
\label{tab:ablation_memory}
\begin{small}
\begin{tabular}{lccccc}
\toprule
Method & PSNR$_{20-25} \uparrow$ & PSNR$_{50-55} \uparrow$ & PSNR$_{70-75} \uparrow$ & PSNR$_{75-80} \uparrow$ & $\Delta$ (Gain) \\
\midrule
w/o GMA     & 21.92 $\pm$ 3.81 & 20.61 $\pm$ 3.75 & 19.98 $\pm$ 3.78 & 19.85 $\pm$ 3.79 & - \\
Random Memory & 15.51 $\pm$ 4.40 & 13.59 $\pm$ 4.47 & 13.34 $\pm$ 4.01 & 13.42 $\pm$ 3.83 & - \\
\midrule
\textbf{W/ GMA} & \cellcolor{red!25}22.83 $\pm$ 3.73 & \cellcolor{red!25}21.65 $\pm$ 3.70 & \cellcolor{red!25}21.04 $\pm$ 3.73 & \cellcolor{red!25}20.92 $\pm$ 3.73 & \cellcolor{red!25}+1.07 \\
\bottomrule
\end{tabular}
\end{small}
\vspace{-0.15in}
\end{table}

Qualitative results confirm that GMA preserves geometric consistency and scene persistence, whereas the w/o GMA baseline suffers from spatial sliding and blurring. While Random memory causes broken geometry, our full model maintains stable structures, validating its effectiveness.

\subsection{Extension}
\begin{figure}[htbp] 
  \centering
  \includegraphics[width=\textwidth]{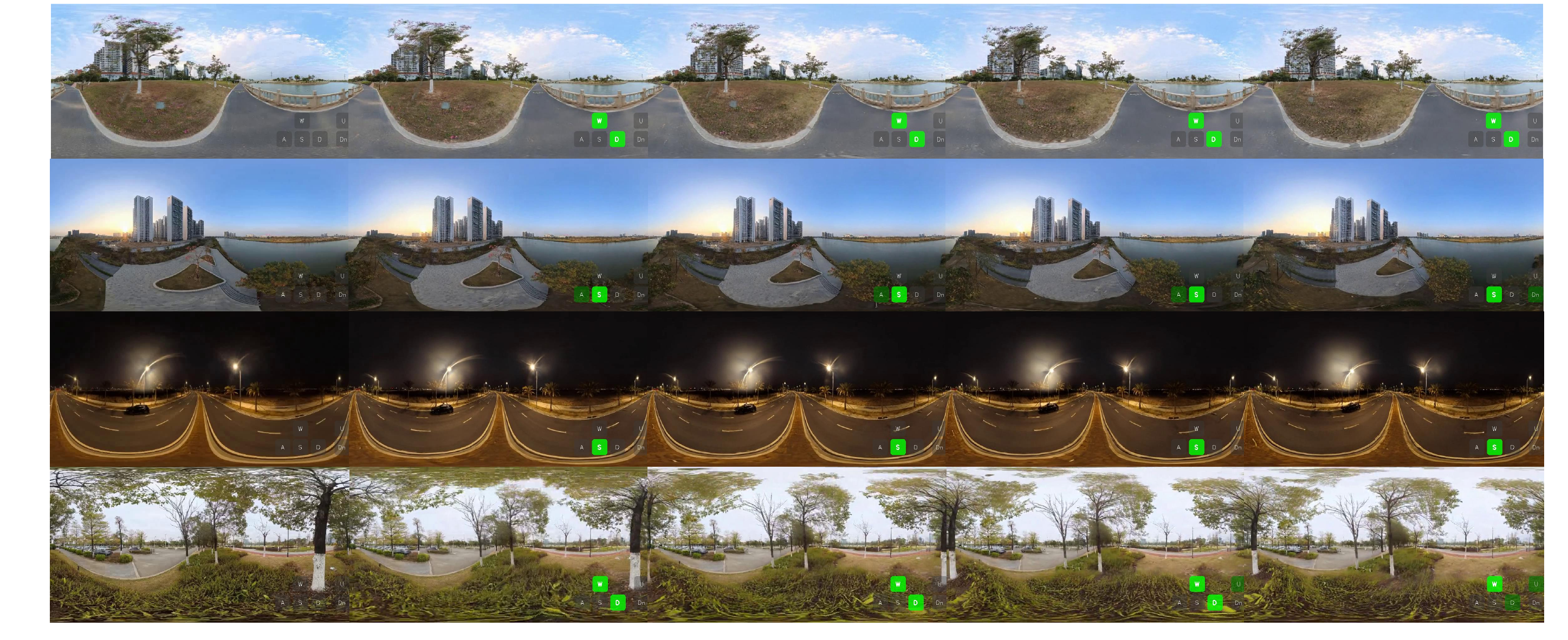} 
  \caption{Real-time Generation. The left column shows the initial input frame, followed by consecutive frames generated in real time based on the specified keyboard inputs.}
  \label{fig:real_time_results} 
\end{figure}

\paragraph{Real-time Generation via Causal Forcing.} We extend our framework to achieve real-time generation via Causal Forcing ~\citep{zhu2026causal}, distilling the full model into a lightweight chunk-wise autoregressive generator. Specifically, we first establish a causal autoregressive framework via teacher forcing, and subsequently match the student generator's distribution $p_{\hat{\theta}}$ to this trained teacher's distribution $p_{\theta}$ by minimizing the asymmetric distribution matching distillation (DMD) loss: $\mathbb{E}_{\hat{x} \sim p_{\hat{\theta}}, x \sim p_{\theta}} [D(\hat{x}_{1:T}, x_{1:T})]$. Benefiting from this distillation and our Rolling Forcing inference, our real-time generator can produce a high-fidelity, 161-frame panoramic video in just \textbf{8 seconds} on a single NVIDIA H20 GPU. This achieves a dramatic, orders-of-magnitude speedup compared to our full model ($\sim$ 4 min 48 s), while being significantly faster than previous state-of-the-art methods such as Matrix-3D ($\sim$ 16.5 min) and OmniRoam ($\sim$ 31 min) with negligible quality trade-offs. Figure~\ref{fig:real_time_results} demonstrates the generation results under this real-time setting. Please refer to the supplementary material for more details.

\section{Conclusion}
In this work, we presented PanoWorld, a framework for controllable panoramic video generation. By leveraging the rotation-equivariance of omnidirectional representations, we decouple rotations to simplify motion learning. Based on the unique properties of panoramic rays, our proposed DPRC and GMA modules ensure rigorous geometric, radiometric, and temporal consistency. To support this research, we introduced World360, a large-scale dataset featuring both real-world UAV captures and high-quality simulations. Extensive experiments on this benchmark demonstrate that PanoWorld significantly outperforms state-of-the-art methods in scene persistence and structural integrity, offering a robust solution for unconstrained panoramic world synthesis.

\begin{figure}[htbp]
  \centering
  \includegraphics[width=1.1\textwidth]{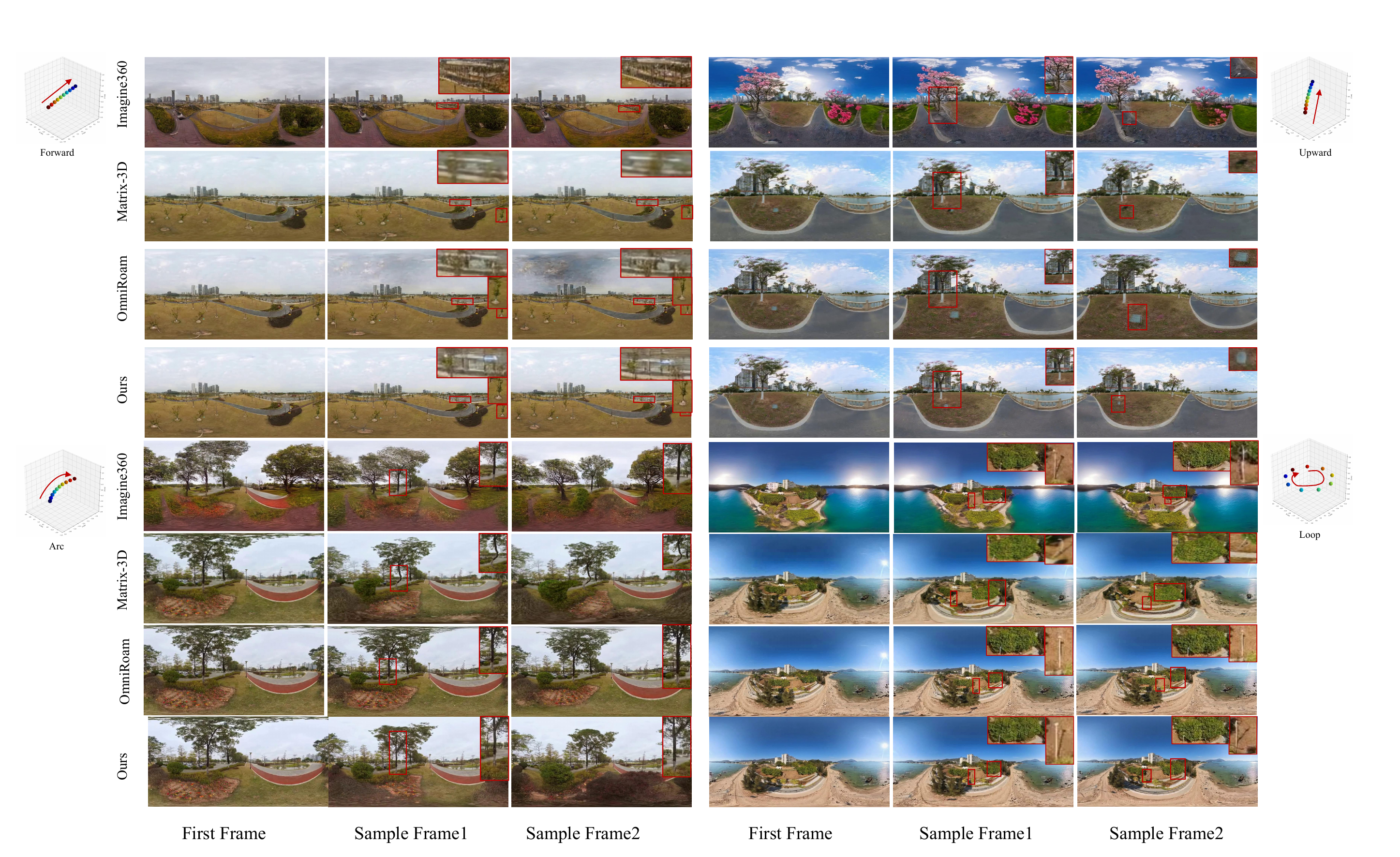} 
  \caption{More results on real-world outdoor sequences.}
  \label{resutls_img} 
\end{figure}

\newpage

\bibliographystyle{plainnat}
\bibliography{reference}

\newpage
\clearpage
\appendix
\begin{center}
    \LARGE Supplementary Material for PanoWorld: Real-World Panoramic Generation
\end{center}
\section{Datasets Pre-processing}
We construct two datasets for model training. The first dataset is \textit{Scene-Reality Data} (referred to as \textit{real-world data}), comprising large-scale panoramic videos captured in diverse physical environments. This dataset exposes the model to natural camera ego-motion and complex visual textures, enhancing its generalization across real-world scenarios. The second is \textit{Trajectory-Precise Data} (referred to as \textit{synthetic data}), generated via AirSim360. By leveraging the high-fidelity simulation and perfect ground-truth trajectories provided by AirSim360, this dataset ensures the model achieves precise, action-conditioned control and maintains high geometric consistency during trajectory execution. 

\subsection{Scene-Reality Real Data}
The Scene-Reality Learning Dataset is a large-scale, synchronized video-inertial dataset captured by a specialized panoramic Unmanned Aerial Vehicle (UAV) in real-world environments. It consists of 124 long-duration 360° panoramic videos and corresponding 6-DoF ego-motion trajectories derived from a high-precision GPS-aided Inertial Navigation System (INS). Totaling over 6 million frames, the dataset covers a wide range of outdoor environments—including parks, roadways, artificial lakes, and architectural complexes—while encompassing highly diverse, unconstrained aerial motion profiles.


\paragraph{Temporal Alignment.}

To ensure strict temporal consistency between the high-rate kinematic data and the panoramic visual stream, we perform a multi-stage synchronization process. This is necessitated by the asynchronous logging of the panoramic camera (59.94 FPS) and the GPS-fused system (200 Hz). The primary objective is to establish a unified timeline by accurately anchoring the video sequence within the continuous pose stream.

In the first step, we perform initial timestamp alignment by matching the camera's exposure start signal with the corresponding entry in the pose sequence to determine the absolute reference time $t_0$. Once the global starting point is anchored, we reconstruct the discrete timeline for the subsequent video frames using the constant frame rate:
\begin{equation}
t_i = t_0 + i \cdot \frac{1}{\text{FPS}},
\end{equation}
where $t_i$ represents the target timestamp for the $i$-th frame. Given that the 200 Hz pose data is recorded on an independent clock, we synchronize the two modalities by matching each calculated $t_i$ to the raw pose sequence via nearest-neighbor search. To mitigate sampling offsets and ensure a smooth trajectory, the discrete pose sequence is resampled to these exact video timestamps using linear interpolation:
\begin{equation}
\mathbf{p}(t_i) = \text{Interp}(t_i, \mathcal{P}_{\text{raw}}),
\end{equation}
where $\mathcal{P}_{\text{raw}}$ denotes the raw trajectory. Finally, any leading video frames that precede the first available pose record are discarded to guarantee that every frame is paired with a valid, synchronized 6-DoF state.

\paragraph{Rotational Decoupling and Coordinate Alignment}
To eliminate violent camera self-rotation and establish a geometrically consistent mapping between the visual content and the kinematic state, we perform a data processing procedure for rotational decoupling and coordinate system alignment. Due to the unconstrained nature of aerial flights, the raw panoramic videos contain significant ego-rotation (e.g., aggressive yawing and wind-induced jitter) that distorts the underlying scene motion. Meanwhile, the raw 6-DoF poses are inherently logged within the Earth-fixed East-North-Up (ENU) coordinate system, which is misaligned with the camera's initial viewing direction. Therefore, our primary objective is to stabilize the visual stream and transform the inertial trajectory into a unified canonical camera coordinate system.

Specifically, we first define the canonical orientation by aligning the ENU coordinate system with the camera coordinate framework of the initial frame. The continuous 3D rotation matrix $\mathbf{R}_t \in SO(3)$ at frame $t$ is then transformed into this aligned canonical frame using the INS metadata. To further refine the alignment and compensate for high-frequency sensor noise, we warp the ERP frames using a hybrid visual-inertial optimization approach. We lift sparse 2D feature correspondences $(u,v)$ from consecutive ERP frames onto a unit sphere $\mathbb{S}^2$ via spherical mapping:
\begin{equation}
\theta = \left(\frac{u}{W} - \frac{1}{2}\right)2\pi, \quad \phi = \left(\frac{1}{2} - \frac{v}{H}\right)\pi,
\end{equation}
\begin{equation}
\mathbf{p} = \begin{bmatrix} \cos\phi\cos\theta \\ \sin\phi \\ \cos\phi\sin\theta \end{bmatrix},
\end{equation}
where $W$ and $H$ denote the ERP width and height. Let $\{\mathbf{p}_i\}$ and $\{\mathbf{q}_i\}$ be the matched 3D unit vectors. The relative rotation $\Delta\mathbf{R}$ is optimized using a RANSAC-based Kabsch alignment initialized by the INS relative orientation:
\begin{equation}
\Delta\mathbf{R}^{\star} = \arg\min_{\Delta\mathbf{R} \in SO(3)} \sum_{i} \|\Delta\mathbf{R}\mathbf{p}_i - \mathbf{q}_i\|_2^2.
\end{equation}
In feature-deficient environments (e.g., open water surface or sky regions), the system automatically falls back to an INS-prior yaw-only estimate combined with latitude-weighted phase correlation along the longitudinal direction. The calculated rotation sequence is smoothed via a one-dimensional constant-velocity Kalman filter. Finally, the rotationally decoupled and stabilized visual stream is generated by applying the inverse rotation matrix to the spherical sampling grid and reprojecting it back to the canonical ERP format via grid remapping, successfully removing camera self-rotation while ensuring that the processed 6-DoF poses are strictly aligned with the canonical video frames.



\paragraph{Pose Noise Suppression}
Due to aerodynamic turbulence, motor vibrations, and occasional GPS multi-path effects during outdoor UAV flights, the raw synchronized 6-DoF poses inevitably suffer from significant high-frequency noise and sudden sensor jitter. To mitigate these artifacts and reconstruct physically plausible, smooth kinematic trajectories suitable for high-fidelity world modeling, we deploy a robust two-stage pose noise suppression strategy. First, a hysteresis filter is applied to the raw discrete pose sequence to eliminate spurious, low-amplitude fluctuations and stabilize the trajectory baseline. Second, to guarantee temporal differentiability and continuous motion, we apply a one-dimensional Gaussian smoothing kernel across the timeline. Considering that vertical positioning typically exhibits higher variance and noise than horizontal tracking, the scale parameters are adaptively configured with standard deviations of $\sigma=0.5$ for the horizontal position components ($x, y$) and $\sigma=1.0$ for the vertical component ($z$), successfully yielding continuous and high-quality trajectory states.

\paragraph{Spatial-Uniform Trajectory Resampling and Visual Alignment}
In real-world data collection, the UAV inevitably flies at variable velocities due to automated path-following adjustments, manual piloting commands, and aerodynamic drag variation. Consequently, trajectories sampled purely on a uniform temporal grid exhibit severe non-uniform spatial distribution, where high-speed segments become spatially sparse and low-speed phases become densely clustered. More importantly, when the drone enters a hovering or stationary state, the temporal sampling continues to log frames despite zero displacement, introducing significant temporal redundancy that contributes nothing to motion-conditioned learning. For downstream generative world models, this spatial non-uniformity and stationary clustering introduce severe motion imbalance and bias the network toward static representations. To resolve these issues, eliminate static frames, and ensure that each discrete data step represents an identical physical displacement, we re-index and resample both the synchronized video and trajectory pairs in the spatial arc-length domain rather than the temporal domain, successfully compressing the non-informative stationary sequences.

The complete engineering pipeline for spatial normalization and sequence generation is implemented through the following structured steps:

\textbf{Step 1: 3D Cumulative Arc-Length Calculation.} 
To successfully map the dataset into the spatial domain, we must first establish a continuous, monotonically increasing spatial metric that reflects the true total distance traversed by the drone. Since our platform is an omnidirectional panoramic UAV, it undergoes complex 3D aerial dynamics that include extensive vertical climbs and descents along the gravitational axis, meaning that a simplified 2D planar distance would fail to capture the vertical motion and distort the alignment. Therefore, to accurately quantify the physical movement, we explicitly integrate the $z$-component into the metric computation and calculate the cumulative three-dimensional Euclidean arc length $s_i$ from the initial frame to the $i$-th frame as follows:
\begin{equation}
s_i = \sum_{k=1}^{i} \Delta s_k = \sum_{k=1}^{i} \sqrt{(x_k - x_{k-1})^2 + (y_k - y_{k-1})^2 + (z_k - z_{k-1})^2},
\end{equation}
where $\mathbf{P}_i = (x_i, y_i, z_i)^T$ represents the discrete 3D position states, and the boundary condition is initialized as $s_0 = 0$.

\textbf{Step 2: Uniform Spatial Re-indexing and Pose Interpolation.} 
With the continuous metric prepared, our next objective is to generate a normalized trajectory benchmark where the kinematic states are uniformly distributed in space. By filtering out temporal steps where $\Delta s_k \approx 0$, this uniform spatial grid inherently compresses the redundant static segments into a single spatial point, thereby balancing the distribution of dynamic motion increments across the entire dataset. To implement this, we define a target spatial grid with a fixed physical interval of $\Delta s = 0.05$\,m, formulating the discrete sampling points $s'_j$ as:
\begin{equation}
s'_j = j \cdot \Delta s, \quad j \in \left\{0, 1, \dots, \left\lfloor \frac{s_{\max}}{\Delta s} \right\rfloor \right\}.
\end{equation}
We then map all continuous kinematic state variables from the original temporal reference frame onto this newly defined spatial grid via continuous piecewise linear interpolation:
\begin{equation}
f(s'_j) = \text{Interp}(s'_j, \mathcal{S}_{\text{raw}}, f(\mathcal{S}_{\text{raw}})), \quad f \in \{x,y,z,\phi,\theta,\psi\},
\end{equation}
where $\mathcal{S}_{\text{raw}} = \{s_0, s_1, \dots, s_N\}$ represents the raw non-uniform spatial coordinate set, and $(\phi,\theta,\psi)$ denote the Roll, Pitch, and Yaw orientation angles. Note that since the trajectory has already been rotationally decoupled and stabilized in the upstream stage, the Euler angles can be directly interpolated without phase-wrapping artifacts.

\textbf{Step 3: Image Frame Differentiation and Temporal Blending.} 
Once the spatially uniform re-sampling indices and target timestamps are determined from the pose stream, the corresponding visual frames must be aligned to match this new spatial grid. Because the target spatial sampling points do not perfectly coincide with the original discrete video frame exposure times, extracting the nearest raw frames directly would introduce severe discretization artifacts and visual judder. To resolve this and ensure smooth visual continuity, we perform a differentiation and interpolation process on the image sequence based on the retrieved resampling IDs. We apply linear temporal blending between two adjacent original frames that bracket the target spatial timestamp $t$:
\begin{equation}
I(t) = (1-\alpha)I_{\lfloor t \rfloor} + \alpha I_{\lceil t \rceil},
\end{equation}
where $\lfloor t \rfloor$ and $\lceil t \rceil$ denote the indices of the preceding and succeeding raw video frames, and $\alpha = t - \lfloor t \rfloor$ represents the fractional interpolation weight. This step successfully forces the video stream to adapt to the spatial domain, guaranteeing that the physical distance between any two consecutive processed frames is precisely maintained at a constant interval of $0.05$\,m. Robustness for cross-platform deployment is further ensured via automatic codec selection for video encoding.

\textbf{Step 4: Fixed-Length Trajectory Slicing.} 
To satisfy the multi-task training objectives of our framework, the continuous, spatially-uniformized video-trajectory sequences are partitioned into two distinct clip lengths using a sliding window strategy. First, to train the motion-conditioned generation model, we extract standard training samples with a fixed length of $L_1 = 81$ frames. Second, to support the optimization of the geometry-aware memory module, we extract longer sequences with a fixed length of $L_2 = 161$ frames. For a given target sequence length $L \in \{81, 161\}$ and a total resampled trajectory length of $M$, the standard non-overlapping slices are defined as:
\begin{equation}
\mathcal{S}_k = \{ \mathbf{P}_j \mid j = kL, \dots, (k+1)L-1 \}.
\end{equation}
When the final remaining segment at the end of a sequence is shorter than the required window size $L$ (i.e., $M \pmod L \neq 0$), we employ a backward-overlapping strategy to construct the last valid slice. Specifically, the final window is obtained by tracing backward from the absolute end of the sequence to capture a complete, continuous block of preceding resampled states, formulated as:
\begin{equation}
\mathcal{S}_{\text{final}} = \{ \mathbf{P}_j \mid j = M - L, \dots, M - 1 \}.
\end{equation}

\subsection{Trajectory-Precise Synthetic Data}
To complement our real-world dataset and provide pristine, noise-free geometric supervision, we curate a trajectory-precise synthetic dataset sourced from the Air360 simulator across 8 diverse outdoor environments. Unlike real-world capture pipelines that suffer from sensor noise, clock drift, and interpolation residues, the simulation platform inherently guarantees perfect, hardware-level spatiotemporal synchronization between the panoramic visual frames and the ground-truth 6-DoF poses. Leveraging this synchronized framework, we develop an efficient, high-throughput automated data collection engine designed to achieve strictly spatial-uniform data acquisition. 

To implement efficient 3D trajectory planning while preventing the camera agent from getting trapped in structurally fragmented spaces or wasting rendering resources on uninformative regions, we devise a hierarchical layer-wise filtering and voxelization strategy. First, the continuous 3D free space of each outdoor scene is discretized into a uniform 3D macro-grid with a resolution of $1\,\text{m} \times 1\,\text{m} \times 1\,\text{m}$. Since complex aerial trajectories naturally involve extensive elevation changes, the efficiency of traditional unconstrained 3D exploration heavily degrades in altitude layers that are overly cluttered or lack traversable continuity. To resolve this and maximize the throughput of the collection engine, our pipeline pre-calculates the exact number of traversable grid cells for each discrete altitude layer. If the count of valid candidate grids within a specific altitude layer falls below a predefined density threshold, indicating insufficient space for meaningful aerial maneuvers, the engine \textbf{automatically bypasses the entire altitude layer}. This proactive layer-wise filtering forces the system to allocate computation exclusively to open, geometrically rich elevations.

For the selected valid altitude layers, candidate anchor points are populated on a dense sub-grid with a refined resolution of $0.05\,\text{m}$ to ensure precise spatial-uniform sampling. To realistically mimic real-world aerodynamic flight statistics and maximize trajectory diversity, the engine is programmed to operate under three complementary trajectory generation modes:

\begin{itemize}
    \item \textbf{Mode 1: Multi-Directional Homogeneous Maneuver.} To establish highly structured multi-view illumination and motion consistency from a unified initial state, the agent departs from a fixed initial waypoint $\mathbf{P}_{\text{start}}$ and executes flight paths along three distinct directional headings, exploring a diverse repertoire of motion profiles (including straight lines, horizontal/vertical curves, and continuous 3D arcs). Each directional flight continues until an environmental obstacle within the $1\,\text{m}$ grid framework is detected, at which point the agent transitions into a stationary hovering state. Crucially, to filter out trivial fragments and ensure sufficient spatial scale for learning long-range dynamics, a multi-directional sequence is considered valid and logged into the corpus if and only if the trajectories generated along all three directional branches concurrently exceed a predefined threshold length $\mathcal{L}_{\text{min}}$.
    \item \textbf{Mode 2: High-Coverage Autonomous Exploration.} To encapsulate intricate 3D maneuvers and maximize environmental familiarity across complex topography, the agent executes active scene coverage from a randomized spatial initialization $\mathbf{P}_{\text{rand}}$. Utilizing a 3D A* algorithm optimized over the $1\,\text{m}$ macro-grid map, the agent dynamically calculates collision-free paths to explore the environment. The trajectory generation process automatically terminates once the cumulative visibility volume covers at least 70\% of the total traversable free space within the current scene, ensuring massive spatial diversity.
    \item \textbf{Mode 3: Randomized Multi-Goal Navigation.} To simulate unconstrained point-to-point transit and capture generic 6-DoF flight statistics, the engine randomly samples both the initial waypoint $\mathbf{P}_{\text{rand\_start}}$ and the target destination $\mathbf{P}_{\text{rand\_end}}$ across the unoccluded 3D grid space. A globally optimal, collision-free trajectory connecting these two randomized terminal states is then generated using the 3D A* pathfinder.
\end{itemize}

To better approximate the smooth, continuous pose transitions observed in real aerial flights and eliminate sharp velocity discontinuities, the piecewise linear edges generated between consecutive anchor points in Mode 2 and Mode 3 are modified using a geometric smoothing procedure. Near each anchor point, adjacent linear segments are replaced with continuous circular arcs constructed to be tangent to the edge vectors while rigorously validating against 3D collision constraints. Since the vicinity spheres of adjacent anchor points may overlap, the tangent offset for each circular arc is limited to 0.3 times the corresponding edge length, ensuring continuous velocity profiles and realistic 6-DoF angular transitions. Finally, continuous panoramic video sequences are rendered along these normalized trajectories at a precise, constant spatial increment of $\Delta s = 0.05\,\text{m}$ directly from the underlying 3D scene representations, yielding high-fidelity, strictly spatial-uniform benchmarks with complex elevation changes and rich geometric constraints.




\subsection{Prompt Construction and Filtering Pipeline}
\label{sec:prompt_filtering}
To enable robust text-conditioned control and multi-modal alignment within the generative world model, the processed video-trajectory sequences must be paired with high-quality, semantically dense textual descriptions. Rather than relying on motion-centric captions that introduce cross-modal redundancy with the explicit 6-DoF trajectory inputs, we construct an automated textual supervision pipeline that focuses exclusively on environmental characteristics, spatial layouts, and ambient illumination. This pipeline bridges stationary scene aesthetics with textual conditions through the following structured operations:

\textbf{Step 1: Environment-Focused Scene Captioning.} 
To synthesize a comprehensive textual representation without introducing computational redundancy from dense frame processing, we uniformly extract a sparse sequence of 21 frames from each video clip. These frames are sequentially fed into a vision-language model (Qwen3-VL) with a tailored systemic prompt that commands the model to describe the background environment while strictly omitting any description of camera ego-motion or dynamic object movements. By observing the multi-frame sequence simultaneously, the model successfully distills stationary scene semantics, structural attributes, weathering conditions, and material textures into a single, cohesive environmental narrative.

\textbf{Step 2: Dual-Level Environmental Granularity.} 
To support different scales of textual conditioning and facilitate both fast semantic anchoring and intricate scene learning, the environment-focused descriptions are adaptively split into a dual-level framework based on word count. Descriptions with a length strictly below 35 words are categorized as \textit{Concise Prompts}, which capture essential high-level scene semantics such as global terrain types (e.g., dense pine forests, rocky canyons) and macro-climatic profiles. Conversely, descriptions exceeding 35 words are preserved as \textit{Detailed Prompts}, providing dense details regarding spatial geometry, architectural styles, fine-grained surface textures, and complex ambient light interactions (e.g., volumetric sun shafts, diffuse overcast shadowing). This multi-granularity design allows the downstream world model to optimize for both coarse-grained environment matching and fine-grained visual-text correspondences simultaneously.

\textbf{Step 3: Multi-Dimensional Illumination and Exposure Assessment.} 
To prevent the generative network from learning corrupted or physically implausible appearance priors from degraded data, the visual frames undergo an automated illumination screening phase across three quantitative criteria based on global pixel intensity statistics:
\begin{itemize}
    \item \textit{Under-Exposure Detection:} To identify severe shadow noise and completely uninformative black regions, we compute the frame-level average pixel intensity $\mu_I$ across the RGB channels. Frames exhibiting a mean brightness value below a strict lower bound, $\mu_I < 30$ (dead dark), are flagged.
    \item \textit{Low-Contrast Tracking:} To capture washed-out scenes with compressed dynamic range, we evaluate the image contrast $\sigma_I$ via pixel intensity standard deviation. Sequences displaying low contrast ($\sigma_I < 20$) are flagged, which typically indicates poorly lit environments when coupled with low mean intensity.
    \item \textit{Over-Exposure Ratio Estimation:} To filter out clipped highlights where geometric textures are entirely lost, we calculate the over-exposure pixel ratio $R_{\text{over}}$. An individual pixel is considered saturated if its intensity exceeds 250, and a video clip is marked as over-exposed if the saturated region occupies a massive portion of the frame, formulated as $R_{\text{over}} > 40\%$.
\end{itemize}.
The identifiers of all video sequences failing any of these multi-dimensional illumination criteria are systematically logged into an exposure anomaly catalog.

\textbf{Step 4: Perceptual Quality Scoring.} 
Beyond traditional pixel-level brightness statistics, generative models are highly sensitive to high-frequency visual artifacts, blurriness, and compression distortions. To quantify human-aligned perceptual quality, we employ Q-Align, a learned blind image quality assessment (BIQA) model, to score the remaining frames. The model predicts a continuous aesthetic and fidelity score $\mathcal{Q} \in [0, 1]$ for each frame based on human perceptual alignment, and the inferred score distribution for the entire video dataset is archived in a scoring directory.

\textbf{Step 5: Strategic Dual-Stage Filtering.} 
To balance processing efficiency with dataset purity across different training scales, our pipeline provides two alternative filtering schemes to generate the clean subset:
\begin{itemize}
    \item \textit{Exposure-Only Filtering (Fast Mode):} This strategy targets rapid data throughput by exclusively pruning the sequences documented in the exposure anomaly catalog (\texttt{exposure\_or\_dark.txt}). It is highly optimized for scenarios requiring quick dataset scaling while ensuring basic illumination consistency.
    \item \textit{Dual-Attribute Filtering (Complete Mode):} To achieve maximum visual purity, this comprehensive strategy executes a joint constraint. It simultaneously discards all exposure-compromised sequences and prunes low-fidelity clips by enforcing a strict perceptual quality cut-off, formulated as $\mathcal{Q} < 0.7$. This dual-gate screening successfully removes all visually degraded content, isolating the highest-quality core subset containing $M$ pristine video sequences.
\end{itemize}

\textbf{Step 6: Dataset Partitioning.} 
With the purified dataset established, our final objective is to partition the corpus into distinct training and testing benchmarks (\texttt{train.csv} and \texttt{test.csv}). We adopt a dynamic dataset splitting strategy where the testing set is assigned a dynamic ratio ranging from 10\% to 20\% of the clean dataset, with the remaining majority allocated to the training set. Crucially, the selection priority for the testing set is strictly bound to video clips that possess comprehensive, high-accuracy ground-truth pose annotations, safeguarding the evaluation phase against trajectory tracking errors and establishing a solid evaluation foundation.





\section{Experiments Settings}
\subsection{Video Model Training Pipeline}
We adopt WAN2.2-5B as our foundational visual generator and integrate Low-Rank Adaptation (LoRA) with a rank configuration of 64 for parameter-efficient fine-tuning. The training is conducted directly utilizing the DiffSynth-Studio framework across a scale of 70k video-trajectory samples. To ensure structural alignment across diverse data streams, all visual sequences are uniformly resized to a spatial resolution of $480 \times 960$. Following the official implementation of WAN2.2-5B, the initial frame of the sequence is concatenated directly with the latent noise along the frame dimension to serve as a strong spatial conditioning anchor, which is subsequently excluded during loss computation to focus the optimization gradient purely on synthesized content. To enhance the model's geometric robustness across multi-modal inputs, we apply random online yaw-angle rotations to the video streams and randomly toggle between concise and detailed environmental prompts. The network is optimized for exactly 2 epochs, with the latitude-longitude loss enabled to strictly enforce spherical geometric consistency. To accommodate hardware memory constraints while maintaining a stable optimization trajectory, the batch size is configured to 1, coupled with a gradient accumulation step of 4, and the complete distributed training pipeline is deployed on 8 NVIDIA H20 GPUs.
\subsection{Action Model Training Pipeline}
To preserve the established multi-modal visual representations and eliminate gradient interference during kinematic alignment, the upstream video generative model is completely frozen, and the action control model is optimized independently. The training protocol is structured into three progressive stages, with each stage executed for exactly 2 epochs to ensure balanced convergence across different data distributions. During the first stage, we initialize the network with basic directional priors by training on the filtered planar line dataset; to artificially expand the spatial coverage and prevent overfitting to specific flight headings, we implement a structured data augmentation scheme by explicitly rotating both the visual frames and their corresponding trajectory yaw angles by fixed orthogonal intervals of $90^\circ$, $-90^\circ$, $180^\circ$, and $-180^\circ$, using a learning rate of $1 \times 10^{-4}$. In the second stage, the training shifts to the entire corpus of 70k real-world data samples to adapt the network to complex, unconstrained physical flight dynamics, maintaining a learning rate of $1 \times 10^{-4}$. In the third stage, the final fine-tuning is executed over 50k trajectory-precise synthetic data samples to further refine the 6-DoF trajectory control limits with a conservatively scaled-down learning rate of $5 \times 10^{-5}$ to safeguard against catastrophic forgetting of real-world features. Throughout both the second and third stages, an aggressive online data augmentation strategy is applied where the video sequences and their paired 6-DoF poses are subjected to synchronized random rotational perturbations to maximize data diversity and 6-DoF angular generalization. For all three training stages, the batch size is strictly configured to 1, coupled with a gradient accumulation step of 4 to maintain an effective optimization batch size while adhering to hardware memory constraints.
\subsection{Memory Model Training Pipeline} 
To maintain the integrity of the established visual generation and action control capabilities, both the upstream video model and the action model are completely frozen, leaving the geometry-aware memory module to be optimized independently. The training protocol is structured into a progressive two-stage scheme. In the first stage, we execute a geometric warm-up phase utilizing 20k frame pairs extracted from the collected video sequences alongside their corresponding ground-truth camera poses. Specifically, the model is fed a complete context frame and a subsequent target frame where a regional mask has been applied. Driven by the relative displacement between the two frames' 6-DoF poses, the memory module is optimized to reconstruct the obscured visual content within the masked region based on the contextual cues from the initial frame. To force the network to learn fundamental, non-trivial spatial transformations rather than local interpolations, the spatial displacement between the two frames in each pair is constrained to strictly within the range of $0.2$\,m to $2.0$\,m.

 In the second stage, we transition to full-sequence learning via block-wise video data loading, utilizing 50k video clips with a fixed length of 161 frames. For each clip, a start frame is randomly sampled from the sequence. To construct the memory conditioning, we greedily select the earliest four frames whose spatial displacements from the sampled start frame strictly lie within the $[0.2, 2.0]$\,m range. These four retrieved frames are then concatenated to form the memory anchor. During training, both this memory anchor and the original conditioning frame are corrupted with noise to supervise the prediction of the subsequent 81-frame video sequence, ensuring sufficient geometric parallax for memory-guided synthesis. Throughout the training process, the learning rate is maintained at $1 \times 10^{-4}$ with a batch size of 1 and a gradient accumulation step of 4 to ensure stable convergence within hardware constraints.

\subsection{Baseline Methods} 

\subsection{Inference} 
During the inference and benchmarking phase, we sample 90 representative video-trajectory sequences from the World360 dataset to serve as a comprehensive evaluation baseline. This evaluation corpus covers a diverse taxonomy of 3D maneuvers, including forward, backward, lateral, vertical (up/down) translations, continuous arcs, and complex loop trajectories. To rigorously quantify model performance, we establish a multi-dimensional evaluation suite spanning distribution fidelity, omnidirectional visual quality, and 6-DoF motion controllability:

\begin{itemize}
    \item \textit{Visual and Geometric Fidelity Metrics:} We employ the standard Fr\'{e}chet Video Distance (FVD) and Fr\'{e}chet Image Distance (FID), alongside panoramic regional variants ($\text{FID}_{\text{pole}}$ and $\text{FID}_{\text{equ}}$), to accurately assess structural integrity and projection consistency across the omnidirectional spherical manifold. To address the inherent geometric distortion introduced by equirectangular projections, visual clarity and fine-grained aesthetic coherence are quantified via the Natural Image Quality Evaluator (NIQE) and learned quality metrics ($\text{QA}_{\text{qual.}}$ and $\text{QA}_{\text{aes.}}$ sourced from Q-Align), which are specifically adapted to account for sphere-to-plane area distortion by applying latitude-aware pixel weighting. Furthermore, temporal consistency and motion-blur artifacts are penalized using the Fr\'{e}chet Audio-Visual Distance (FAED) equivalent adaptations, establishing an evaluation framework inherently tailormade for panoramic video synthesis rather than standard planar configurations.
    \item \textit{Trajectory Controllability and 6-DoF Adherence:} To guarantee a deterministic and fair comparison across different generative baselines, we condition all competing methods on identical ground-truth (GT) trajectories. This uniform conditioning effectively eliminates evaluation biases induced by varying motion scales or high-frequency tracking noise typically generated by conventional Structure-from-Motion (SfM) pipelines. We compute the average Peak Signal-to-Noise Ratio (PSNR) between the synthesized videos and the ground-truth sequences over multiple progressive temporal windows to quantify structural persistence and alignment under explicit motion constraints. 
    
    Crucially, since pixel-level metrics like PSNR cannot fully capture absolute geometric drifting in 3D space, we introduce a more stringent, trajectory-focused validation mechanism using \textbf{ViPE}, a state-of-the-art visual pose estimation network optimized for ego-motion tracking. We utilize ViPE to explicitly reconstruct the continuous 6-DoF camera trajectories directly from the generated panoramic video frames. The recovered trajectories are then aligned and compared against the reference ground-truth poses using standard robotic trajectory metrics, such as Absolute Trajectory Error (ATE) and Relative Pose Error (RPE). This inverted geometric reconstruction provides an explicit, highly sensitive gauge of the model's action-following accuracy and physical consistency over long-horizon rollouts.
\end{itemize}



\section{Application}
\subsection{Real-Time Application via Causal-Forcing}
\paragraph{Qualitative Results.} We present additional qualitative results for real-time  video generation utilizing our developed panoramic I2V world model. In contrast to previous methods that suffer from severe structural blurring or motion inconsistencies due to architectural gaps, our framework—built upon Causal Forcing and advanced to the Wan2.2-TI2V-5B backbone—demonstrates superior capabilities. Given a single input panoramic image and a sequence of camera trajectories, our model successfully generates temporally coherent, high-fidelity panoramic scene videos. Thanks to the chunk-wise teacher forcing and causal ODE distillation, the generated previews exhibit exceptionally stable scene geometry, precise perspective transitions, and strict adherence to the specified camera motion over extended temporal horizons.
\paragraph{Technical Implementation}
To build a high-fidelity, controllable interactive environment simulator, we adapt the Wan2.2-TI2V-5B diffusion model, transforming its original dual-stream architecture into a causal autoregressive framework via Causal Forcing.

We implement a chunk-wise autoregressive strategy that partitions the 21 latent frames into a $[1] + [4 \times 5]$ structure. The initial frame serves as the clean I2V conditioning anchor and is excluded from loss computation. The remaining 20 frames are trained with teacher forcing in 4-frame chunks. To preserve the dual-stream nature of the backbone while enforcing causality, the full ground-truth (GT) latent sequence is concatenated with the noisy sequence to form the input ($clean_x$), and an I2V-specific teacher forcing attention mask is constructed via FlexAttention. This mask guarantees that: (i) the noisy tokens of the conditioning frame only attend to their own clean counterparts, and (ii) the noisy tokens within a specific generation chunk attend solely to the clean GT tokens from preceding chunks, preventing future information leakage within the same chunk. Camera poses are duplicated along the token dimension to precisely align with this clean/noisy dual-stream sequence structure.

 We follow the standard three-stage pipeline of Causal Forcing to compress the autoregressive Wan2.2 model into an efficient real-time generator: (1) autoregressive diffusion training with teacher forcing, (2) causal ODE initialization to bridge the architectural gap, and (3) asymmetric distribution matching distillation (DMD). At 480p resolution, Stage-1 is trained for 3,000 steps, and Stage-3 DMD is optimized for 10,500 steps, yielding an ultra-fast 4-step generator with discretized timesteps at $\{1000, 750, 500, 250\}$.

 \textbf{Long-video distillation \& Rolling Forcing Inference.} To transcend the 21-latent-frame (81 pixel frames) training limit, we extend chunkwise DMD to 161-pixel videos (41 latent frames) via a Rolling Forcing mechanism. During training, rollouts autoregressively generate up to 41 causal latent frames using a sliding-window KV cache, while DMD supervision is applied to the last 21 frames for compatibility with frozen bidirectional score networks. We distill the model on 1,998 pre-encoded panoramic clips. At inference, Rolling Forcing generates all 41 latent frames in a single pass conditioned on the first frame, text, and camera trajectories (synthetic or dataset poses), yielding 161-frame videos with a 4-step distilled scheduler. To eliminate contamination of the clean cache within the moving denoising windows, a prefix snapshot-and-restore mechanism for the KV caches is utilized. Furthermore, for memory efficiency during long-sequence generation, chunked VAE decoding with a 1-latent-frame (corresponding to 4 pixel frames) overlap blending mechanism is implemented.

\textbf{Quantitative Results.} We report the quantitative evaluation on panoramic video generation and motion controllability in Table~\ref{tab:main_results}. As demonstrated, our full model (\textbf{Ours}) significantly outperforms existing state-of-the-art methods, achieving the best scores across almost all visual quality (e.g., FID, FAED), perceptual quality (QA), and trajectory controllability (PSNR) metrics. 

\begin{figure}[t]
\centering
\includegraphics[width=\textwidth]{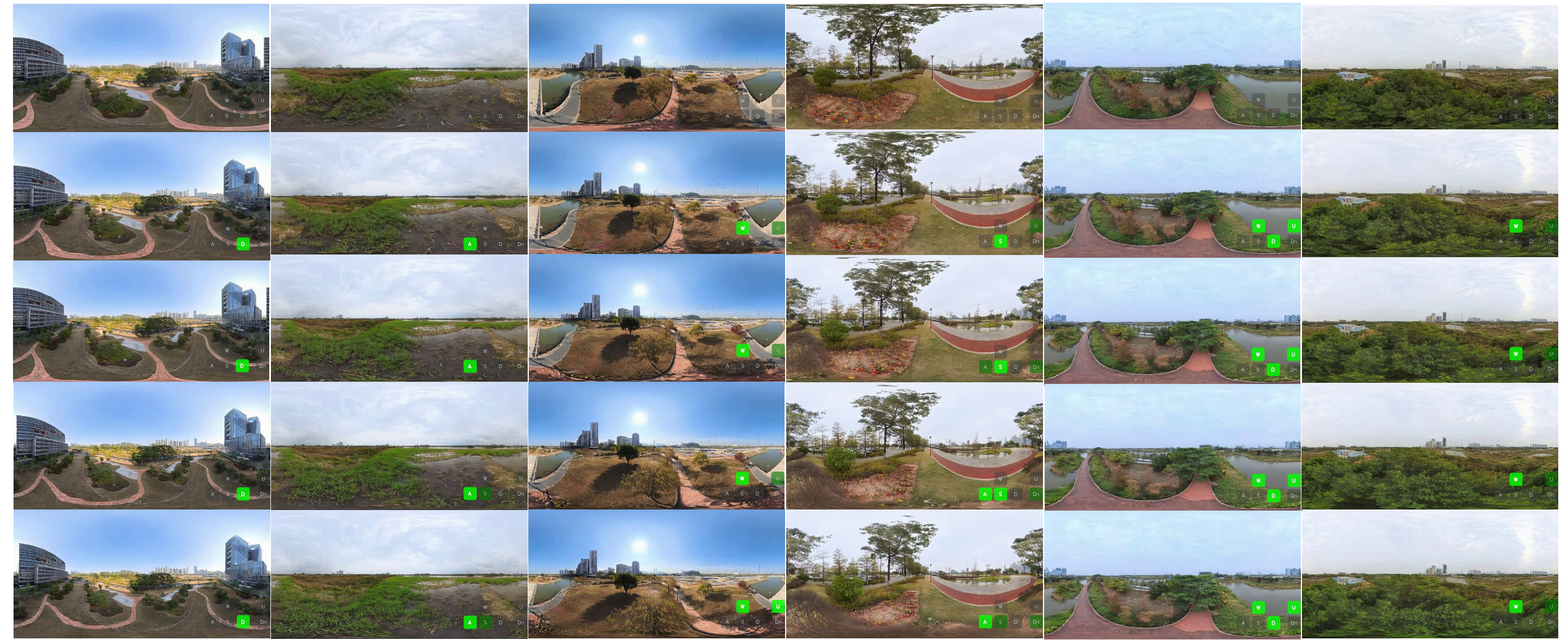}
\caption{Comparison of Generation Efficiency}
\label{Generation Efficiency}
\end{figure}

When introducing the causal-forcing variant, it exhibits only a marginal and acceptable drop in both visual fidelity and trajectory control compared to our full model. However, this negligible quality gap is accompanied by a dramatic, orders-of-magnitude improvement in inference efficiency. As illustrated in Figure~\ref{Generation Efficiency}, on a single NVIDIA H20 GPU, our full model requires 4 minutes and 48 seconds to complete the generation, whereas competitive methods like Matrix-3D and OmniRoam suffer from prohibitive computational costs, taking up to nearly 16.5 and 31 minutes, respectively. In stark contrast, our causal-forcing model successfully slashes the generation time to a mere 8 seconds. This exceptional speedup and the clear efficiency trade-off shown in Figure~\ref{Generation Efficiency} establish causal-forcing as a highly practical and powerful solution for real-time generation.

\begin{table}[h]
\centering
\caption{Quantitative results on panoramic video generation and motion controllability.}
\label{tab:main_results}
\setlength{\tabcolsep}{3pt}
\begin{small}
\begin{tabular}{l c c c c c c c c c c}
\toprule
\multirow{2}{*}{Method} & \multirow{2}{*}{ Gen. Time}& 
\multicolumn{5}{c}{Visual Quality} &
\multicolumn{2}{c}{Perceptual Quality} &
\multicolumn{2}{c}{Traj. Controllability} \\
\cmidrule(r){3-7} \cmidrule(lr){8-9} \cmidrule(l){10-11}
& & FID$\downarrow$ & FID$_p$$\downarrow$ & FID$_e$$\downarrow$ & FAED$\downarrow$ & NIQE$\downarrow$ & QA$_q$$\uparrow$& QA$_a$$\uparrow$& PSNR$_{25}$$\uparrow$ & PSNR$_{55}$$\uparrow$ \\
\midrule
Matrix-3D & 16m36s & 34.63 & 67.88 & 68.64 & 3.39 & 4.74 & 2.99 & 2.13 & 20.47 & 18.80 \\
OmniRoam & 30m46s & 60.77 & 85.25 & 45.92 & 3.36 & 3.39 & 3.75 & 3.06 & 15.51 & 13.59 \\
\quad self-forcing & 16s & 107.32 & 153.35 & 112.98 & 3.79 & 3.73 & 2.57 & 2.26 & 15.32 & 13.62 \\
\textbf{Ours} & 4m48s & 27.64 & 47.21 & 26.00 & 2.63 & 3.85 & 4.02 & 3.128 & 22.83 & 21.65 \\
\quad causal-forcing & 8s & 28.07 & 56.66 & 27.27 & 2.81 & 4.06 & 3.96 & 3.09 & 21.93 & 20.32 \\
\bottomrule
\end{tabular}
\end{small}
\end{table}

\section{Discussion}
\subsection{Limitation}
\label{limitation}
Despite its efficiency, our current framework exhibits certain limitations. The primary bottleneck stems from the strategy of directly replicating the input image as the first frame of the generated sequence. While this approach bypasses the denoising cost of the initial frame to accelerate inference, it introduces a noticeable distribution discrepancy between the high-fidelity real input and the subsequent model-generated frames. Consequently, a distinct quality gap emerges starting from the second frame, which gradually accumulates and degrades temporal consistency over extended horizons. Furthermore, although the introduced memory mechanism effectively mitigates this degradation, the overall synthesis remains heavily dependent on the first frame, potentially constraining the model's capacity to handle extensive environmental drift or major structural transitions.
\subsection{Future Work}
\label{future_work}
To address the aforementioned limitations, we envision several promising directions for future exploration. 
First, to bridge the domain gap caused by directly injecting the input image, we plan to investigate lightweight boundary-refinement modules or latents-space alignment techniques. Instead of hard replication, optimizing the transition from the real input to the generated latent sequence could significantly preserve visual fidelity in subsequent frames. 
Second, to mitigate the heavy dependency on the initial frame and improve long-term temporal stability, we aim to develop a dynamic memory management mechanism. By continuously updating the memory pool with newly generated, high-confidence frames and introducing active forgetting strategies, the model can adaptively refresh its scene representation. This would reduce error accumulation and better support navigation through highly dynamic, large-scale environments over extended horizons.

\newpage

\newpage
\end{document}